\documentclass[pdflatex,sn-mathphys-num]{sn-jnl}

\usepackage{graphicx}%
\usepackage{multirow}%
\usepackage{amsmath,amssymb,amsfonts}%
\usepackage{amsthm}%
\usepackage{mathrsfs}%
\usepackage[title]{appendix}%
\usepackage{xcolor}%
\usepackage{textcomp}%
\usepackage{manyfoot}%
\usepackage{booktabs}%
\usepackage{algorithm}%
\usepackage{algorithmicx}%
\usepackage{algpseudocode}%
\usepackage{listings}%

\usepackage{acronym}
\usepackage{siunitx}
\usepackage{lineno}

\acrodef{UAV}{uncrewed aerial vehicle}
\newacroindefinite{UAV}{an}{a}
\acrodef{AID}{Avian-informed drone}
\newacroindefinite{AID}{an}{a}
\acrodef{DoF}{degrees of freedom}
\newacroindefinite{DoF}{an}{a}
\acrodef{BO}{Bayesian optimization}
\newacroindefinite{BO}{an}{a}
\acrodef{AoA}{angle of attack}
\newacroindefinite{AoA}{an}{a}
\acrodef{AoS}{angle of side slip}
\newacroindefinite{AoS}{an}{a}
\acrodef{INDI}{Incremental non-linear Dynamic Inversion}
\newacroindefinite{INDI}{an}{a}
\acrodef{MPC}{Model Predictive Control}
\newacroindefinite{MPC}{an}{a}
\acrodef{PID}{Proportional--Integral--Derivative}
\newacroindefinite{RL}{an}{a}
\acrodef{IMU}{Inertial measurement unit}
\newacroindefinite{IMU}{an}{a}
\acrodef{EKF}{Extended Kalman Filter}
\newacroindefinite{EKF}{an}{a}
\acrodef{ROS}{Robotic Operating System}
\newacroindefinite{ROS}{an}{a}

\begin{document}

\title[Article Title]{Adaptive morphing of wing and tail for stable, resilient, and energy-efficient flight of avian-informed drones}


\author*[1]{\fnm{Simon L.} \sur{Jeger}}\email{simon.jeger@epfl.ch}
\author[1]{\fnm{Valentin} \sur{W{\"u}est}}\email{valentin.wueest@epfl.ch}
\author[1]{\fnm{Charbel} \sur{Toumieh}}\email{charbel.toumieh@epfl.ch}
\author[1]{\fnm{Dario} \sur{Floreano}}\email{dario.floreano@epfl.ch}

\affil*[1]{\orgdiv{Department of Mechanical Engineering}, \orgname{Laboratory of Intelligent Systems}, \orgaddress{\street{EPFL}, \city{Lausanne}, \postcode{1005}, \state{Vaud}, \country{Switerland}}}

\abstract{Avian-informed drones feature morphing wing and tail surfaces, enhancing agility and adaptability in flight. Despite their large potential, realising their full capabilities remains challenging due to the lack of generalized control strategies accommodating their large degrees of freedom and cross-coupling effects between their control surfaces. Here we propose a new body-rate controller for avian-informed drones that uses all available actuators to control the motion of the drone. The method exhibits robustness against physical perturbations, turbulent airflow, and even loss of certain actuators mid-flight. Furthermore, wing and tail morphing is leveraged to enhance energy efficiency at \qtyproduct{8}{m/s}, \qtyproduct{10}{m/s} and \qtyproduct{12}{m/s} using in-flight Bayesian optimization. The resulting morphing configurations yield significant gains across all three speeds of up to \qtyproduct{11.5}{\%} compared to non-morphing configurations and display a strong resemblance to avian flight at different speeds. This research lays the groundwork for the development of autonomous avian-informed drones that operate under diverse wind conditions, emphasizing the role of morphing in improving energy efficiency.}



\maketitle

\section{Introduction}\label{sec1}

\acp{AID}~\cite{grant2010flight,di2017bioinspired, xu2019morphing, ajanic2020bioinspired, chang2020soft, ajanic2022sharp, zhang2022, brody2023matagull} are endowed with morphing wing and tail surfaces that make them more agile and capable of flying at a larger range of speeds and with lower energy consumption than fixed-wing drones with comparable mass and aerial surface~\cite{bowman2002evaluating, jha2004morphing, bowman2007development, mintchev2016adaptive, harvey2021aerodynamic, harvey2022review}.
However, their complex design and high number of controllable degrees of freedom pose a challenge for control systems due to the coupling effects of actuators (e.g., wing sweep affects roll, pitch, and yaw rates), state dependencies (e.g., the aerodynamic effect of wing sweep increases with a higher angle of attack), and interactions between joint actuators (e.g., the aerodynamic effect of wing sweep depends on the wing twist angle). In addition, unsteady dynamic effects, such as airflow through the feathers, aeroelasticity~\cite{van2020passive}, and asymmetries arising from imperfections in the manufacturing process, increase the difficulty of developing accurate models.
Consequently, existing \acp{AID} are typically remotely operated by human pilots who cannot take full advantage of the large number of independently controlled \ac{DoF}.

Human teleoperation is made possible by combining multiple actuators into a single control command or by predefining a set of discrete configurations that can be selected by the operator during flight~\cite{di2017bioinspired, ajanic2020bioinspired, ajanic2022sharp}. Other approaches support the pilot with closed-loop control on the elevator to maintain pitch during flight~\cite{chang2020soft}. Recent works on variable sweep \acp{UAV} have used cascaded \ac{PID} control~\cite{greatwood2017perched,waldock2018learning} and Reinforcement Learning~\cite{fletcher2021reinforcement} to increase pitch agility in perching manoeuvres. However, these strategies have not been applied in more general flight scenarios. Recently~\cite{liu2023employing}, proposed a nonlinear dynamics model that combined conventional ailerons with asymmetric wing sweeps at small \ac{AoA} $\in [0^{\circ}, 8^{\circ}]$; this strategy achieved faster roll rates but neglected the influence of wing sweep on other body rates, thus not fully leveraging their potential. In summary, no existing control strategy systematically uses all \ac{DoF} to control motion and therefore leverages the full potential of \acp{AID}.

Here, we propose a method for \ac{AID}-control in which a dynamics model is employed to derive a mapping between the desired body rates and control surface deflection, leveraging all available \ac{DoF} and accounting for coupling effects, state dependencies, and interdependent actuation.
This mapping enables the use of a closed-loop control system, combining model knowledge with the ability to tune gains on the system, making the approach robust to modelling discrepancies.

We validate the applicability of the proposed method on an \ac{AID} with morphing wing and tail surfaces~\cite{ajanic2022sharp}, showing stable flight under physical perturbations, turbulent airflow, and even the loss of certain actuators. Furthermore, \acp{UAV} with morphing aerial surfaces can produce stable flight in several different wing and tail configurations. We show that exploring these configurations in a sample-efficient way significantly increases energy efficiency at speeds of \qtyproduct{8}{m/s}, \qtyproduct{10}{m/s} and \qtyproduct{12}{m/s}. The resulting morphing pattern is analysed and compared to that of birds of similar size and mass.
The results and methods described in this article provide a foundation for the development of autonomous \acp{AID} capable of following trajectories in diverse wind conditions and performing energy-efficient flights across a wide range of operations.

\section{Results}\label{sec2}
\subsection{Stable flight despite disturbances}
\begin{figure}[h!]
    \centering
    \includegraphics[width=\textwidth]{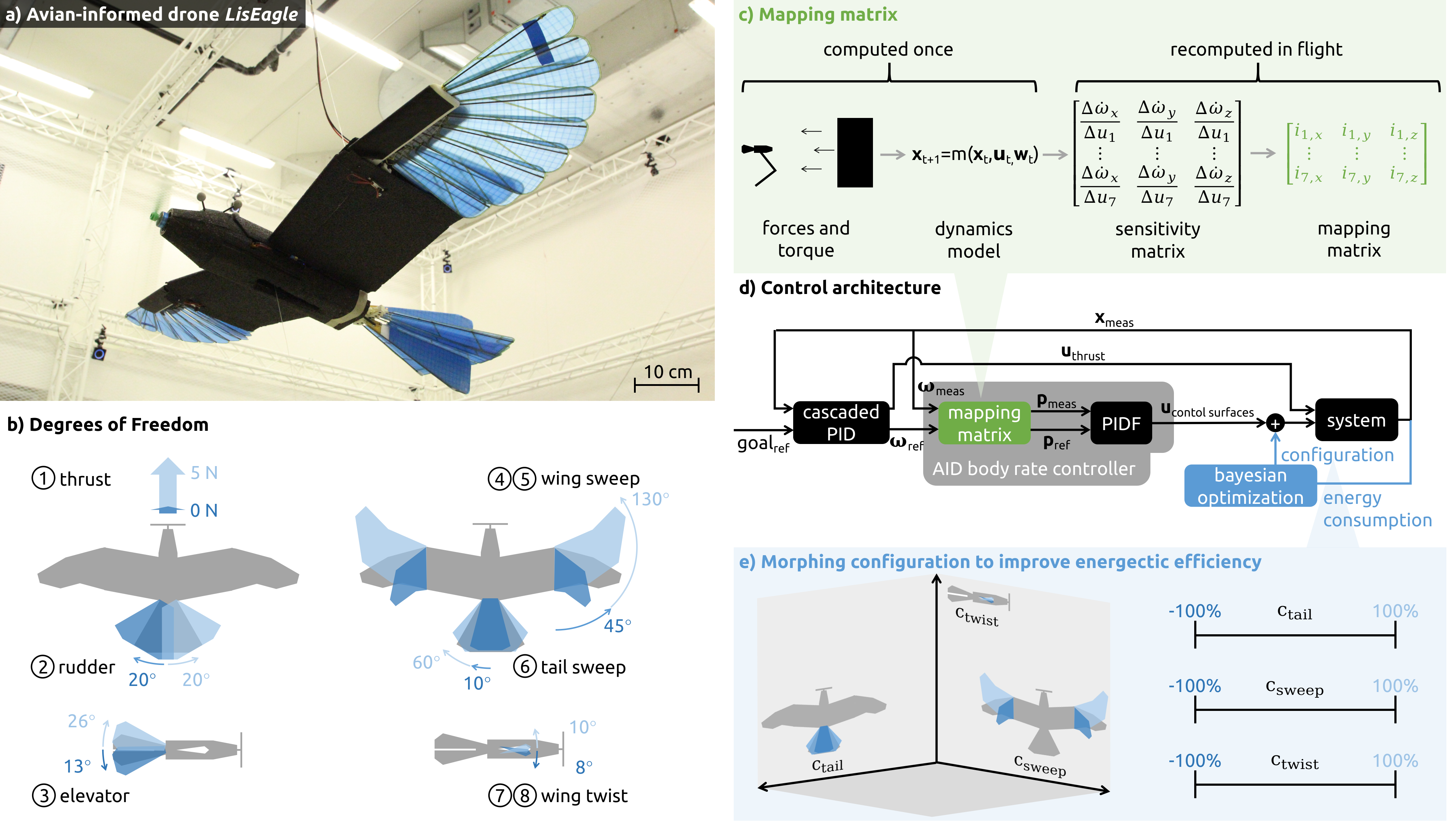}
    \caption{\textbf{Overview}. \textbf{a} The avian-informed drone \textit{LisEagle} in free flight, with a wing span of \qtyproduct{1.52}{m} and a ready-to-fly mass of \qtyproduct{752}{g}. \textbf{b} It features eight degrees of freedom, with interdependent actuation between the wing sweep and twist and between the tail sweep and elevator position. \textbf{c} Wind tunnel experiments determine aerodynamic coefficients at various actuator positions, angle of attack, and angle of side slip. The interpolated data formulates a dynamics model, enabling the computation of the sensitivity matrix which yields a mapping matrix, that scales the importance $i$ of each control surface proportional to its influence on the body rate. \textbf{d} A cascaded control architecture produces reference body rates and thrust, with the latter directly applied to the drone. The mapping matrix (green) projects measured and reference body rates into the actuator space ($p_{\text{meas}}, p_{\text{ref}}$), enabling the use of Proportional--Integral--Derivative control with a feedforward term $F$. \textbf{e} An additional control loop (blue) using Bayesian optimization to minimize energy consumption. The configuration of the drone, described by $c_{\text{tail}}, c_{\text{sweep}}, c_{\text{twist}}$, modifies the centre positions of the tail sweep, symmetric wing sweep, and symmetric wing twist, influencing the lift-to-drag ratio and hence energetic consumption.
    }
    \label{fig:method}
\end{figure}

The  \textit{LisEagle}~\cite{ajanic2022sharp}, an avian-informed drone with eight \ac{DoF} (\autoref{fig:method}a, b), was flown in an environment with constant airflow generated by our indoor setup (Method \ref{sec:setup}).
Using a dynamics model (Method \ref{sec:model}) grounded through wind tunnel measurements, a state-dependent mapping (Method \ref{sec:mapping}) is established between the desired body rates and actuator deflection (\autoref{fig:method}c). The control command (Method \ref{sec:control}) is then adjusted using a tuneable feedforward coefficient $F$, and a \ac{PID} loop is applied to reduce the remaining body rate error (\autoref{fig:method}d).
The resulting body rate controller is embedded within a cascaded control structure, enabling stable flight against constant airflow. In this work, flow speeds of \qtyproduct{8}{m/s}, \qtyproduct{10}{m/s} and \qtyproduct{12}{m/s} were chosen to mimic typical flight velocities for fixed-wing \acp{UAV} of similar size and mass to the \textit{LisEagle}~\cite{stastny2019flying}. Identical PIDF-gains and a control frequency of \qtyproduct{50}{Hz} are maintained across all experiments, matching the actuation frequency of the servos.

\begin{figure}
    \centering
    \includegraphics[width=0.6\textwidth]{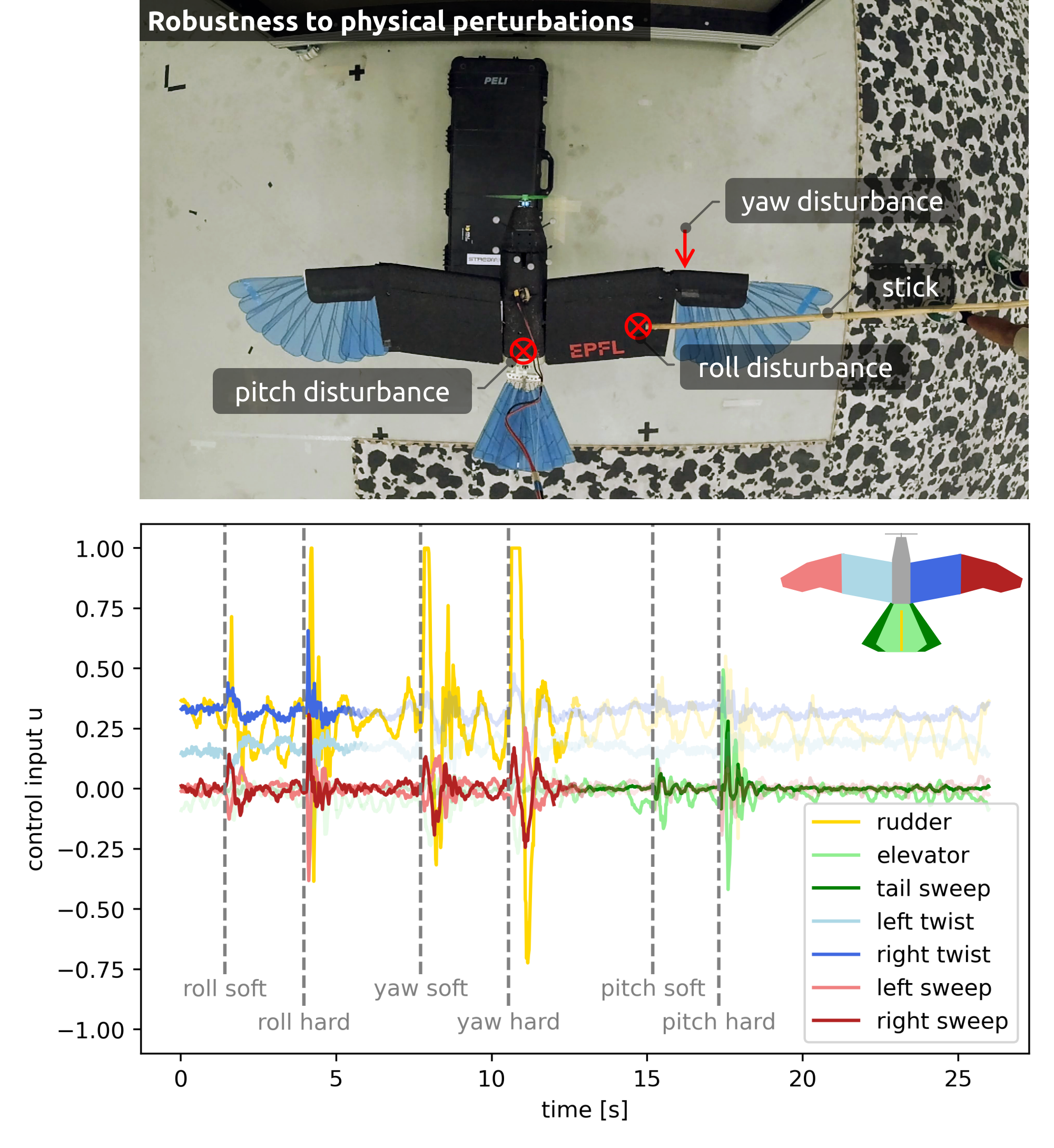}
    \caption{\textbf{Control commands respond to physical perturbations in flight.} The \textit{LisEagle} is disturbed at its extremities using a rod to induce torque in the roll, yaw, and pitch axis. Perturbations are categorized as soft or hard based on the force applied during the disturbance. To enhance readability, control commands are depicted transparently for scenarios in which their impact is minimal.}
    \label{fig:poke}
\end{figure}

To study resilience to external perturbations, the \textit{LisEagle} is manually disturbed in flight with a rod to induce torques around the roll, pitch, and yaw axis (Supplementary Material, Video \ref{video:perturbation}).
The flight data (\autoref{fig:poke}) shows that perturbations around the roll axis are mainly counteracted by wing twist and sweep, which, together with the rudder, help address adverse yaw effects. For perturbations around the yaw axis, the rudder plays the main role, aided by asymmetric wing sweep. To address perturbations in the pitch direction, the controller predominantly deflects the elevator and increases tail sweep.
These results show that the proposed method leverages all available \ac{DoF} to stabilize the drone.
It should be noted that this behaviour would be different for high \ac{AoA} values since the influence of the actuators is state-dependent (e.g., the aerodynamic effect of wing sweep compared to wing twist increases with a higher \ac{AoA}), which is accounted for by the mapping matrix.

\begin{figure}
    \centering
    \includegraphics[width=0.6\textwidth]{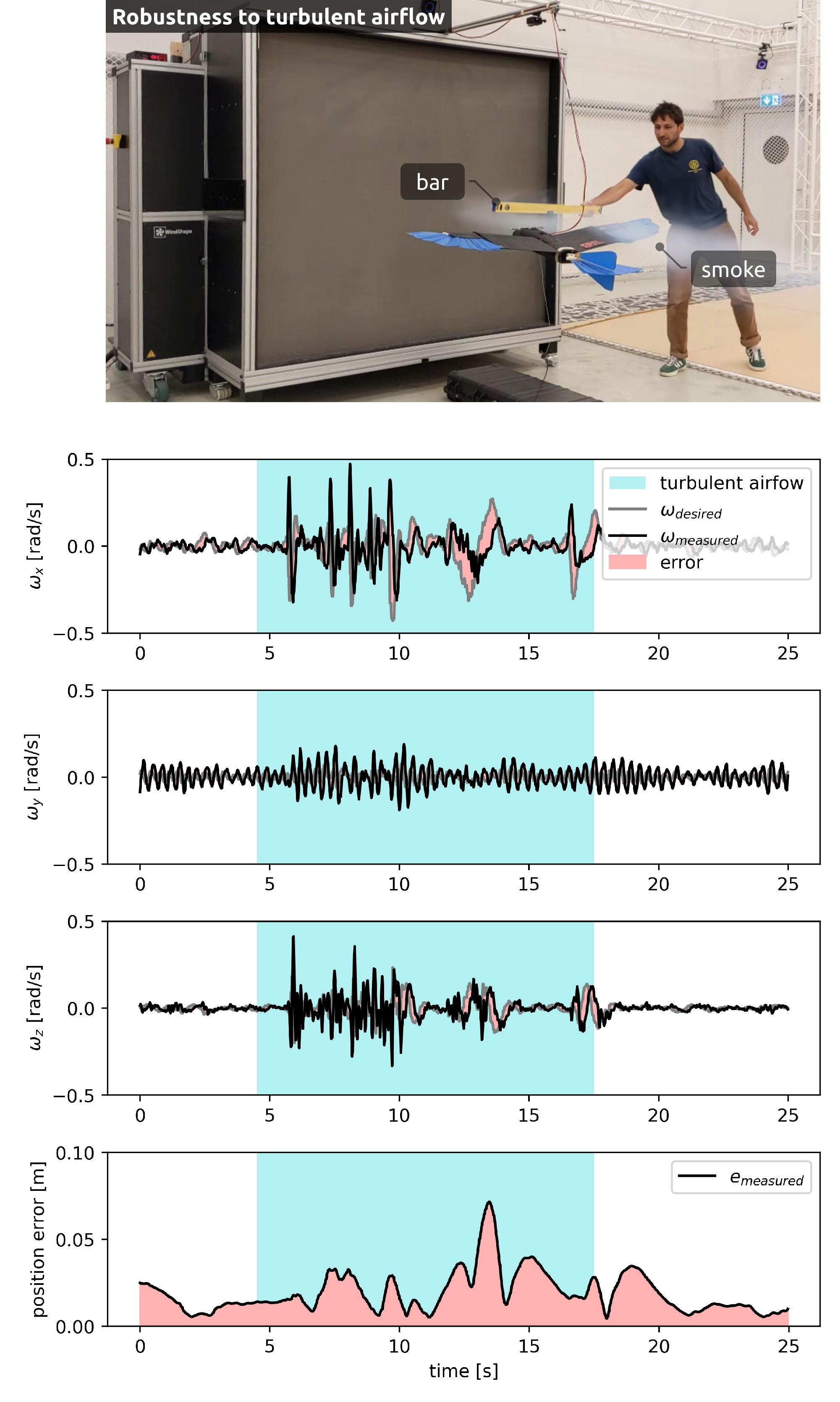}
    \caption{\textbf{Controller is robust to turbulent airflow.} Turbulence is induced over the right wing of the \textit{LisEagle} by using a bar to asymmetrically disturb the airflow, visualized through white smoke.}
    \label{fig:smoke}
\end{figure}

To study the behaviour of the \ac{AID} in turbulent air, the airflow over the right wing of the \textit{LisEagle} is actively disturbed in flight (Supplementary Material, Video \ref{video:turbulence}).
The flight data (\autoref{fig:smoke}) shows large changes in the roll and yaw rates while the pitch rate is less affected. This is expected, as the airflow is asymmetrically disturbed in the lateral direction. The resulting position error increases to \qtyproduct{7}{cm} but quickly recovers after the disturbance stops and returns to the initial accuracy of \qtyproduct{2}{cm}, thus highlighting the robustness of the proposed method.
Note that in this experiment, a scenario is created in which the airflow is asymmetrically disturbed but still primarily comes from one uniform direction. This is a first step towards achieving flight in challenging wind conditions; however, the reality of naturally occurring turbulent, gusty winds from different directions has not yet been considered.

\begin{figure}
    \centering
    \includegraphics[width=0.6\textwidth]{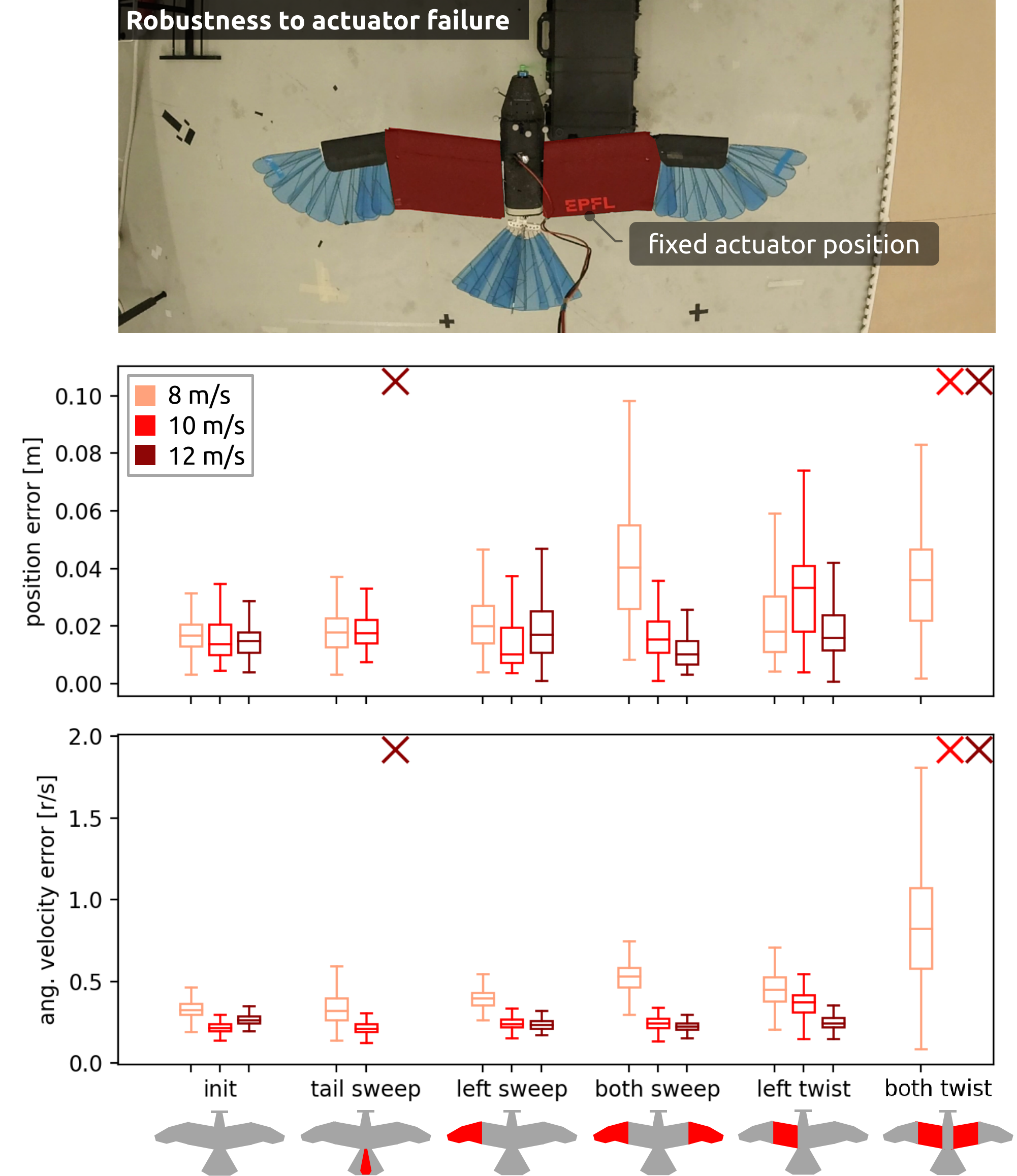}
    \caption{\textbf{Actuator loss is simulated by fixing specific control commands during flight.} Position and angular velocity errors are analysed, with each column representing an experiment conducted at three different speeds throughout \qtyproduct{20}{s}. Unstable behaviour is denoted by a cross.}
    \label{fig:defect}
\end{figure}

The avian-informed LisEagle is an overactuated system: Elevator and tail sweep both influence pitch, wing sweep and twist are mainly responsible for roll. This enables the compensation of actuator failure. To study the resilience of the control method to loss of actuation authority, we immobilize the actuator positions in mid-flight: Wing sweep and twist are fixed to their midpoint position, while tail sweep, due to the mechanical pretension in the system, is set to the furled position. These experiments were conducted at flight speeds of \qtyproduct{8}{m/s}, \qtyproduct{10}{m/s}, and \qtyproduct{12}{m/s} (Supplementary Material, Videos \ref{video:actuator8}, \ref{video:actuator10}, \ref{video:actuator12}), with actuators blocked while the drone was airborne. The flight data (\autoref{fig:defect}) show that the \textit{LisEagle} remained in stable flight when losing the ability to control wing sweep on one or both sides and when losing control of wing twist on one side for all three flight speeds. Flight without both-sided wing twist is possible only at \qtyproduct{8}{m/s} because low speeds lead to an increase in \ac{AoA}, where wing sweep is more effective for roll control than wing twist. The sudden change in tail size when furling the tail sweep leads to unstable behaviour at \qtyproduct{12}{m/s} because changes in the lift surface have greater effects at higher speeds~\cite{beard2012small}. Overall, these results show that the proposed method can leverage to redundant nature of the drone to compensate for unexpected mechanical failures in flight.

\subsection{Morphing can increase energy efficiency}
As shown in the previous experiment, the over-actuated nature of the \textit{LisEagle} enables steady flight with different wing and tail configurations. This creates a null space that can be changed in mid-flight (\autoref{fig:method}e).
Here we define a configuration as three parameters ($c_{\text{tail}}, c_{\text{sweep}}, and c_{\text{twist}} \in [-1,1]$), that is added to the control command $\boldsymbol{u}$ (\autoref{eq:conf}). This changes the centre position of the tail sweep, symmetric wing sweep and symmetric wing twist, which affect the lift-to-drag ratio and consequently the energy consumption of the drone.

\begin{alignat}{2}
    \label{eq:conf}
    u_{\text{tail sweep}} &\leftarrow u_{\text{tail sweep}} \cdot (1 + c_{\text{tail}}) &&\in [15^\circ, 60^\circ] \nonumber \\
    u_{\text{left sweep}} &\leftarrow u_{\text{left sweep}} \cdot (1 + c_{\text{sweep}}) &&\in [50^\circ, 125^\circ] \nonumber \\
    u_{\text{right sweep}} &\leftarrow u_{\text{right sweep}} \cdot (1 + c_{\text{sweep}}) &&\in [50^\circ, 125^\circ] \\
    u_{\text{left twist}} &\leftarrow u_{\text{left twist}} \cdot (1 + c_{\text{twist}}) &&\in [-8^\circ, 10^\circ] \nonumber \\
    u_{\text{right twist}} &\leftarrow u_{\text{right twist}} \cdot (1 + c_{\text{twist}}) &&\in [-8^\circ, 10^\circ] \nonumber
\end{alignat}

\begin{figure}
    \centering
    \includegraphics[width=\textwidth]{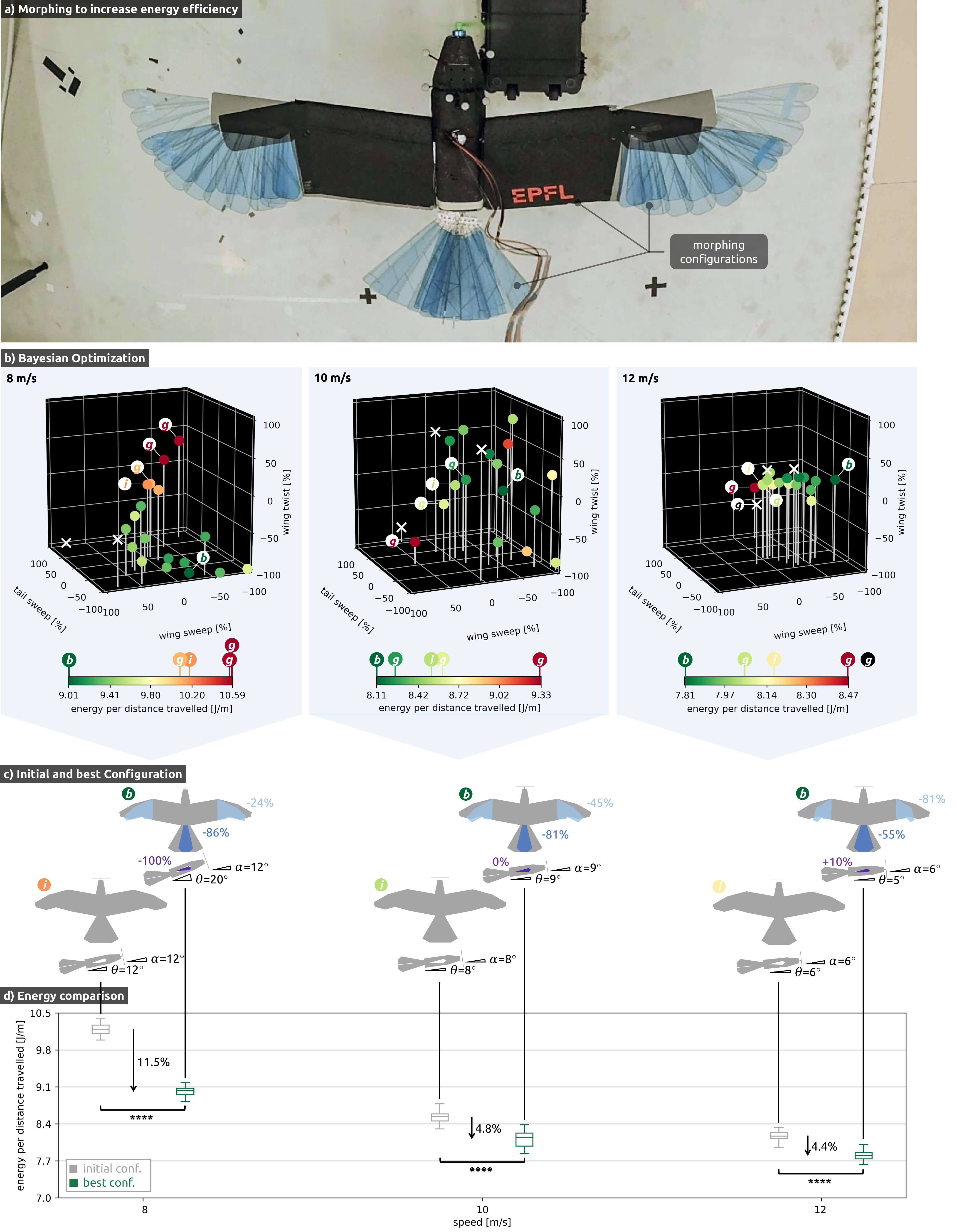}
    \caption{\textbf{Leveraging morphing to increase energy efficiency.} \textbf{a} Overlay of different morphing configurations in flight. \textbf{b} Visualization of the \ac{BO} algorithm exploring $20$ different configurations (coloured dots) per flight. Each experiment starts with the initial configuration $i$ followed by three model-based guesses $g$ to initialize the \ac{BO}, which determines the next $16$ configurations. The best configuration $b$ corresponds to the lowest measured energy consumption at each velocity. The colour of each dot represents the respective energy consumption measurement divided by the airspeed (energy consumed per distance flown). The white crosses indicate unstable configurations (angular velocities $ > $ \qtyproduct{0.5}{rad/s} or position error $ > $ \qtyproduct{13}{cm}). \textbf{c} Illustration of the initial and best configurations and their resulting angle of attack of the wing $\alpha$ and pitch angles $\Theta$. \textbf{d} Comparison of the energy efficiency between the initial and best configurations at \qtyproduct{8}{m/s}, \qtyproduct{10}{m/s} and \qtyproduct{12}{m/s}, measured over \qtyproduct{10}{s} at \qtyproduct{5}{Hz}. The level of significance is indicated by the number of stars.}
    \label{fig:energy}
\end{figure}

To study to what extent wing and tail morphing can decrease energy consumption at different speeds, we use \ac{BO} (Method \ref{sec:bo}) to search for the most energetically efficient configurations (\autoref{fig:energy}a) while the drone is in flight at \qtyproduct{8}{m/s}, \qtyproduct{10}{m/s}, and \qtyproduct{12}{m/s} (Supplementary Material, Videos \ref{video:energy8}, \ref{video:energy10}, \ref{video:energy12}).

The flight data (\autoref{fig:energy}b) show that there are multiple configurations for a given speed that yield nearly identical energetic efficiency, indicating the non-convex nature of the problem. The \ac{BO} process identifies configurations for each speed with lower energy consumption than the initial model-based best-guess configurations, highlighting the importance of acquiring data during flight.
The most efficient configurations (\autoref{fig:energy}c) identified by \ac{BO} reveal that decreasing wing sweep and increasing wing twist lead to more energy-efficient flight at increasing flight speeds. In particular, symmetric wing twist has a large influence on the pitch angle of the drone, enabling stable flight even at \qtyproduct{20}{^\circ} pitch (\autoref{fig:energy}c) which reduces energy consumption at low speeds.
Wing and tail morphing significantly reduce energetic consumption at all flight speeds ($p< 0.0001$, Mann Whitney U test); the energy consumption reduction with respect to the initial configuration is largest at \qtyproduct{8}{m/s} (\qtyproduct{11.5}{\%}) and decreases with increasing flight speed (\qtyproduct{4.4}{\%} at \qtyproduct{12}{m/s}). In addition, the lowest energy consumption was observed for the highest measured flight speed (\qtyproduct{12}{m/s}), suggesting that the corresponding wing and tail configuration is the most suitable for long-range flights. 

\

\section{Discussion}\label{sec3}
The proposed control method maps deflections of morphing actuators into body rates and, unlike other methods, addresses coupling effects, state dependencies and interdependent actuation effects. When applied to an avian-informed drone with 8 controllable degrees of freedom, it produced stable flight (\qtyproduct{2}{cm} precision in constant airflow) and displayed robustness against physical perturbations, disturbed airflow, and control loss of some degrees of freedom. 
Since the method produces control commands in body rates, it can be used in combination with several different high-level controllers, including machine learning approaches, such as Reinforcement Learning, which could leverage it to generate complex manoeuvres~\cite{kaufmann2023champion}, thus opening the door to fully autonomous flight of avian-inspired drones.

Accuracy on the mapping matrix and therefore on control could be further improved by including the inertia of moving actuators and aeroelastic effects induced by the soft feathers in the model.
Since the proposed experimental method enables in-flight data collection for various morphing configurations and flight speeds, it could also be used to collect measurements for data-driven modeling~\cite{selig2010modeling, torrente2021data, rohr2023credible}.

We also showed that the proposed control method can be used to leverage the over-actuated nature of avian-informed drones with multiple degrees of freedom to adapt their wing and tail configurations to different flight speeds for lower energy consumption. The Bayesian Optimization method used in this study requires relatively few sampling points and this can be used to iteratively identify the most efficient configuration for a given flight speed while the drone remains stable in the air. Furthermore, the resulting configurations closely align with behaviours observed in birds of comparable size and mass. The barn owl (Tyto Alba), tawny owl (Strix Aluco) and goshawk (Accipiter gentilis) all reduce their wing spans at higher velocities and decrease the angle of attack of the wing~\cite{cheney2021raptor}. A reduction in the wing area at higher speeds has also been observed for smaller birds, such as swallows~\cite{thomas1996flight} and swifts (Apus apus)~\cite{lentink2007swifts}. 

Assuming that the wind speed is relatively constant, the proposed approach could be used also in outdoor environments because all the required computation, including Bayesian Optimization, run on an onboard companion computer. We speculate that the most energetic configurations identified for each speed (\qtyproduct{8}{m/s}, \qtyproduct{10}{m/s}, \qtyproduct{12}{m/s}) could be interpolated to produce efficient flight at intermediate flight speeds that were not assessed in the experiments. However, outdoor flights would require the integration of sensing technologies for measuring wind (e.g., Pitot Tube, angle of attack sensor) and accurate localization (RTK-GPS or Computer Vision).

We also hypothesize that wing and tail morphing could be used not only to reject external disturbances, as shown in the experiments described here, but also to compensate for other temporary changes in direction and magnitude of the airflow~\cite{quinn2019lovebirds,cheney2020bird}, which were not studied here (e.g., asymmetric wind gusts from below the drone), and even leverage them for reducing energy consumption, as observed in some birds~\cite{laurent2021turbulence,yang2022fast}. 


In conclusion, the methods and results described in this article pave the way for fully autonomous drones with extensively morphing wing and tail surfaces, opening the door to an unmatched combination of agility and adaptability for energy-efficient flight in unexpected and changing conditions.

\section{Method}\label{sec4}
\subsection{Setup}\label{sec:setup}
\begin{figure}
    \centering
    \includegraphics[width=0.6\textwidth]{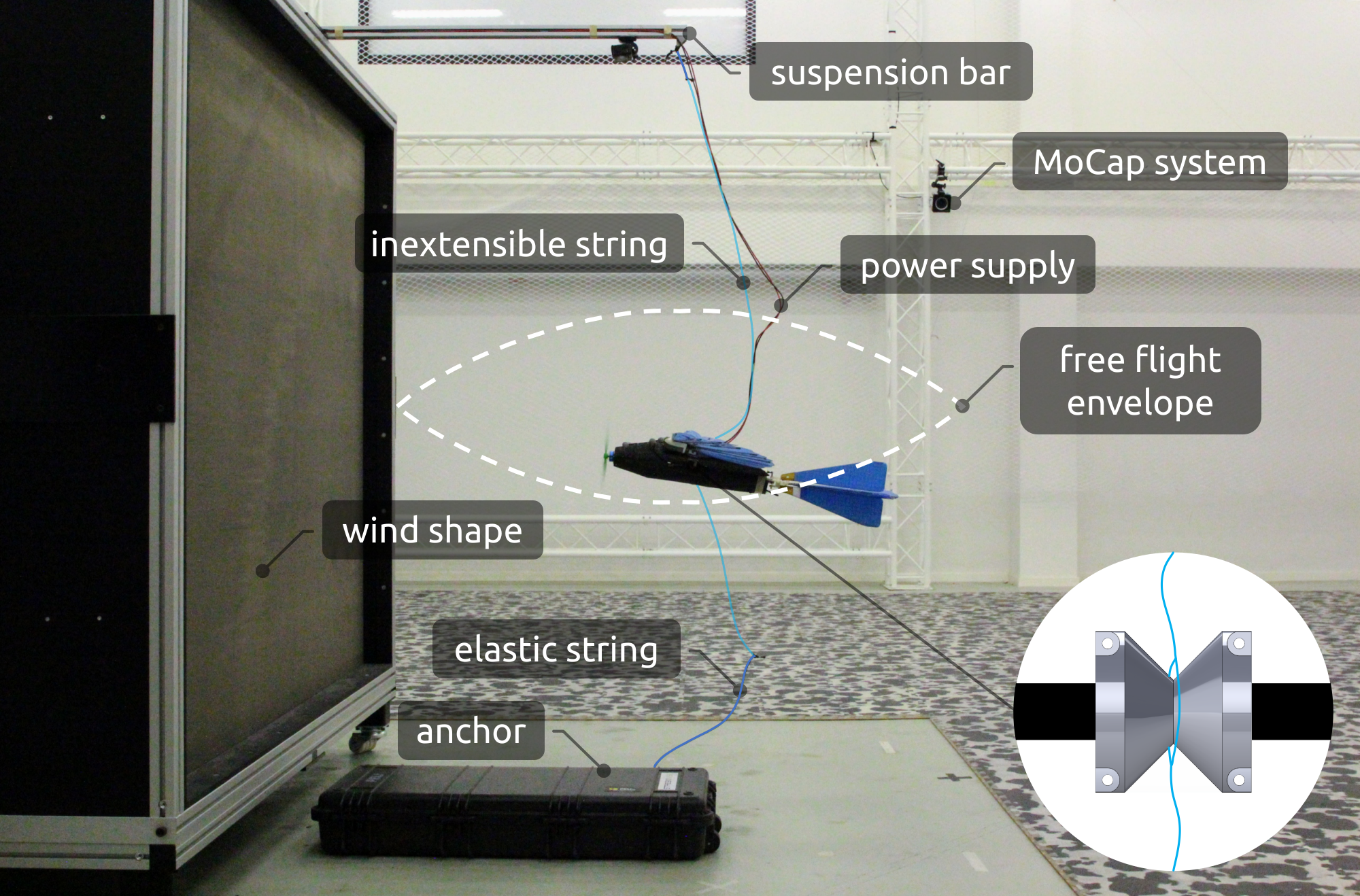}
    \caption{\textbf{Safe testing setup for indoor flight tests.} An open jet wind tunnel produces constant airflow. The strings on the top and bottom include elastic (light blue) and inextensible parts (dark blue), allowing for a free flight envelope of approximately \qtyproduct{40x20x20}{cm} (white dashed line) in which the strings remain slack. The elastic section minimizes the impact of forces in the case of unexpected behaviour. The attachment point to the drone (bottom right) allows for free angular motion, even when the strings are under tension. This configuration was used for gain tuning of the body-rate controller.}
    \label{fig:setup}
\end{figure}

Safe testing is crucial for research on \acp{AID}. Due to their complexity, a crash could lead to a setback of weeks or even months, with the risk of major repairs compromising repeatability. In contrast to rotary wing \acp{UAV} or ground-based robots, returning to a safe state is not trivial for \acp{AID}. To safely conduct flight tests, the \ac{AID} is tethered loosely in front of an open jet wind tunnel~\cite{windshape}, permitting free motion within a fixed volume (\autoref{fig:setup}). Beyond these bounds, the \ac{AID} is held in place by a string that consists of elastic and inextensible parts, limiting the range of motion while also reducing the forces acting on the system in the case of uncontrolled behaviour.

The indoor setup (Supplementary Material, Video \ref{video:takeoff}), similar to that of ~\cite{suys2023autonomous}, offers controlled wind conditions and precise state estimation without drift. The position and orientation are determined via a motion capture system~\cite{optitrack}, transmitted over WiFi through the \ac{ROS}~\cite{ros} to the \textit{LisEagle} and combined with the inertial sensor data through an \ac{EKF}~\cite{ribeiro2004kalman} on a Pixhawk 4 autopilot~\cite{pixhawk4}. All computations related to control are performed on the companion computer, an Nvidia Jetson Nano~\cite{nvidia_jetson_nano} with a custom carrier board~\cite{lis_vision_flight_2023} that communicates directly with Pixhawk 4 to minimize latency.

To support longer testing sessions, a constant voltage power supply is used instead of a battery. The latter is still carried on board to maintain the original weight.
In the experiments performed with this setup, the \textit{LisEagle} has not experienced any major damage, suffering only from fatigue in a small number of servos and feathers due to extensive testing over approximately \qtyproduct{200}{min} of flight time.

\subsection{Dynamics model}\label{sec:model}
Creating a precise dynamics model of an \ac{AID} is difficult due to the unsteady dynamics (such as airflow through feathers and aeroelasticity), interdependent actuation (e.g., the control of the wing sweep depends on the wing twist), and coupling effects (e.g., wing sweep affects pitch, roll, and yaw). Additionally, the presence of inertia-heavy control surfaces that are difficult to model and imperfections caused by the complex manufacturing process can lead to asymmetric behaviour.

Our dynamics model integrates classic fixed-wing theory~\cite{beard2012small}, modified to accommodate morphing effects as described in~\cite{ajanic2020bioinspired}. To address the complexity of the feathered wing structure, aerodynamic coefficients are grounded using wind tunnel measurements. Static aerodynamic coefficients are obtained via the ATI Nano25 force and torque balance~\cite{nano} at a wind speed of \qtyproduct{10}{m/s}. Utilizing a St\"aubli TX-90 robotic arm~\cite{robotarm}, the \ac{AID} is positioned at various \ac{AoA} and \ac{AoS} (in steps of 4\textdegree\ in the range of [-8\textdegree,40\textdegree] and steps of 10\textdegree\ in the range of [40\textdegree,90\textdegree]), executing different actuator combinations. Dynamic aerodynamic coefficients are determined through additional experiments using the forced oscillation method~\cite{vicroy2012static, klein2001estimation}. To reduce the number of measurements, the system is assumed to be symmetric along the longitudinal plane, and the centre of gravity, centre of lift, and moment of inertia are adjusted depending on the wing sweep.
Additionally, control surface movement is modelled as a second-order system fitted to motion capture data, considering limited actuator acceleration, velocity, and delay. The influence of the propeller is modelled using the Tyto Robotics Series 1580 thrust stand~\cite{thrust} as a first-order system with delay.

The proposed model does not account for dynamic effects such as aeroelasticity and the resulting change in airflow through the feathers or the effect of inertia on the actuators. While the latter is unusual for standard \ac{UAV} dynamics models, as control surfaces represent a small portion of the system's mass and thus introduce negligible inertia-based torques, the impact of actuator inertia effects on sensor measurements and the resulting difference between the model predictions and measured accelerations is analysed in Appendix \ref{app1}.

\subsection{Mapping}\label{sec:mapping}
Using the dynamics model, a mapping from the desired body rates to the actuator deflections is created. While this mapping is assumed to be constant for rotary- or fixed-wing \acp{UAV} (e.g., ailerons consistently affect roll~\cite{poksawat2016automatic}), for \acp{AID}, this mapping is state-dependent (e.g., wing twist is more effective for roll at lower \ac{AoA} values than wing sweep, which has a larger influence at higher \ac{AoA} values) and actuator dependent (e.g., wing sweep moves the centre of gravity and centre of lift, influencing the pitch stability of the drone~\cite{ajanic2022sharp}). Additionally, there are interdependencies among certain control surfaces: the effect of the wing sweep is highly dependent on the wing twist angle, and the effect of the tail sweep is a function of the elevator deflection.

To account for these dependencies, the mapping matrix $M$ is recalculated during flight (Algorithm \ref{alg:m}) at a rate of \qtyproduct{10}{Hz}, balancing the computational load and accuracy. The dynamics model $m$ is evaluated at the current state $\boldsymbol{x}$ with actuator positions slightly smaller ($u_i - \Delta u/2$) and larger ($u_i + \Delta u/2$) than the current position $u_i$. Here, $\Delta u/2$ constitutes \qtyproduct{10}{\%} of the full actuator range, reflecting realistic changes within one timestep.
This allows calculating the sensitivity matrix $\boldsymbol{S}$, illustrating the impact of the seven control surfaces on the body rates (with the motor assumed to exert only linear forces). Subsequently, the sensitivity values are normalized column-wise according to the body rate, producing the mapping matrix $\boldsymbol{M}$. This normalization distributes the use of each control surface proportionally to its influence on the body rate (e.g., at low \ac{AoA}, the wing twist has a larger impact on the roll rate than does the wing sweep, which dominates at high \ac{AoA}).

Note that various approaches can be used to create a mapping based on the sensitivity matrix. Calculating the pseudoinverse would result in the smallest change in actuator deflection but can lead to asymmetrical configurations in steady-state flight.
For simplicity, column-wise normalization is chosen, resulting in symmetrical steady-state behaviour.

\begin{algorithm}
\caption{Updating the mapping matrix $\boldsymbol{M}$}
\label{alg:m}
\begin{algorithmic}[1]
\State $i = 0$\;
\For{$i=1 \leq 7$ (all control surfaces)}
    \State $\dot{\boldsymbol{\omega}}_{\text{min}} = m(\boldsymbol{x}, u_i - \Delta u/2)$
    \State $\dot{\boldsymbol{\omega}}_{\text{max}} = m(\boldsymbol{x}, u_i + \Delta u/2)$
    \State $\Delta \dot{\boldsymbol{\omega}} = \dot{\boldsymbol{\omega}}_{\text{max}} - \dot{\boldsymbol{\omega}}_{\text{min}}$
    \State $\boldsymbol{S_i} = \Delta \dot{\boldsymbol{\omega}} / \Delta u$
\EndFor
\For{$c=1 \leq 3$ (all columns)}
    \State $\boldsymbol{M_c} = \boldsymbol{S_c}/\sum_{i=1}^{i \leq 7}{|S_{c,i}|}$ \;
\EndFor
\end{algorithmic}
\end{algorithm}

\begin{figure}
    \centering
    \includegraphics[width=0.42\textwidth]{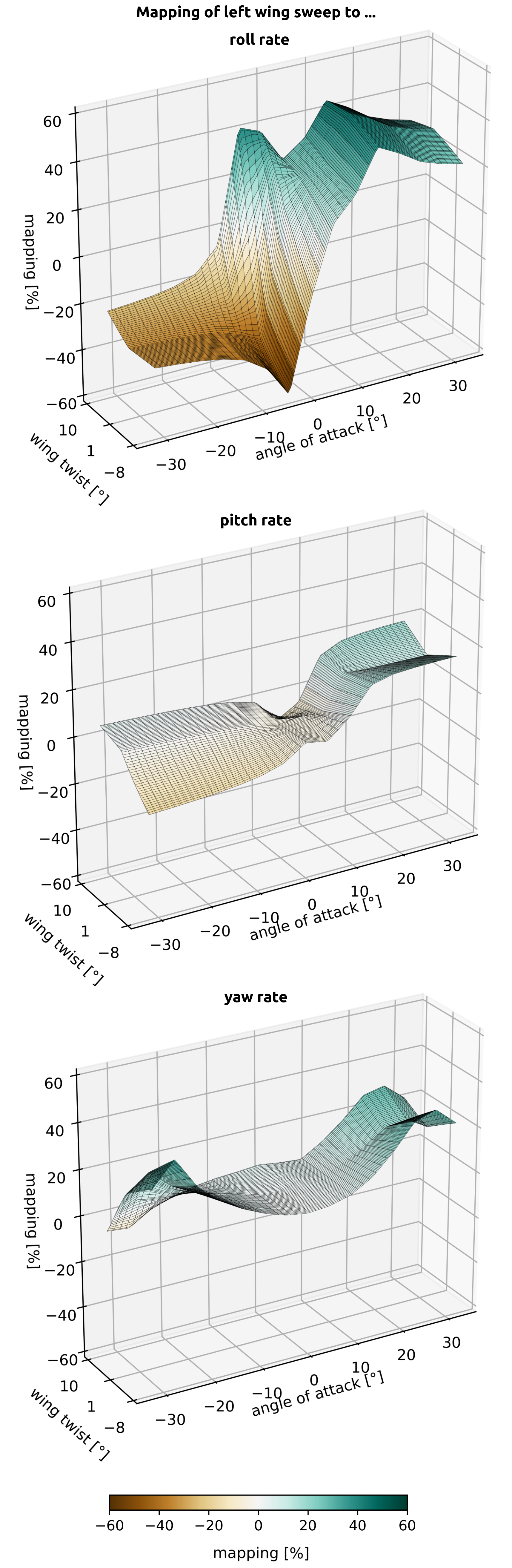}
    \caption{\textbf{The influence of actuators is state-dependent.} The mapping of the wing sweep at \qtyproduct{8}{m/s} as a function of the angle of attack and the wing twist in the roll, pitch, and yaw directions. Each element in the mapping represents the influence of an actuator on the respective body rate (e.g., at high angle of attack, wing sweep has a large impact on the roll rate, so it is used more than at lower angle of attack, where twisting is prioritized). This visualisation represents row $m_\text{left sweep}$, one of seven rows in the mapping matrix (one for each control surface).}
    \label{fig:mapping}
\end{figure}

On the basis of this mapping, the body rates are projected to the actuator space, a necessary step for control (\autoref{fig:method}c): the body rates are split into different components, each corresponding to an actuator (e.g., the desired roll rate is distributed \qtyproduct{15}{\%} to the left sweep, \qtyproduct{10}{\%} to the right sweep, \qtyproduct{40}{\%} to the left twist, and \qtyproduct{35}{\%} to the right twist). This results in a $7 \times 3$ matrix, with each row representing a control surface deflection and each column representing a body rate.
A single row of this mapping matrix is shown in \autoref{fig:mapping}, demonstrating the need for frequent recalculation during flight due to its dependency on the state and actuator position.

This concept is independent of the type of \ac{AID} and the number of \ac{DoF} and can be applied to any \ac{UAV}, given a basic dynamics model.

\subsection{Control}\label{sec:control}
Using the mapping, angular velocity references $\boldsymbol{\omega}_\text{ref}$ and measurements $\boldsymbol{\omega}_\text{meas}$ are projected into the actuator space ($\boldsymbol{p}_\text{ref}$, $\boldsymbol{p}_\text{meas}$), enabling a control scheme to track a reference on each actuator individually (\autoref{fig:method}d).

Different approaches exist to provide closed-loop control with a focus on high robustness against modelling discrepancies. Experiments on rotary-wing~\cite{smeur2016adaptive} and flying wing~\cite{tal2021global} \acp{UAV} using \ac{INDI} show high-performance nonlinear control with strong robustness to modelling inaccuracies. Although a suitable candidate at first glance, it relies on acenergetic efficiencyceleration measurements, which, in the case of \acp{AID}, are altered by the movement of their inertia-heavy control surfaces (\autoref{fig:problem}a). This introduces momentary torques into the sensor readings, making acceleration-based control impractical.
The most common approach for body rate control of \acp{UAV} remains \ac{PID} control due to its low complexity and model-free nature, which enables tuning of the real system parameters~\cite{sobolic2009agile, poksawat2016automatic, poksawat2017gain, bulka2018aerobatics, sattar2020roll, susanto2021application}. This approach can be directly applied to body rates, making it more tolerant to the torques generated by the inertia-heavy actuators of \acp{AID}.

Given the focus on body rate control, a feedforward term, scaled by the coefficient $F$, is incorporated. This component provides a constant actuator deflection (a constant roll rate requires constant control input), serving as a model-based estimate for the control command. The \ac{PID} control strategy addresses the resulting error~\cite{visioli2004new}. This setup mirrors a PIDF controlling a single integrator system, stabilized in a closed loop after gain tuning~\cite{nise2020control}.

Furthermore, the gain scheduling strategy, as outlined in Algorithm \ref{alg:co}, incorporates the influence of the airspeed $v$ on the properties of the control surface (e.g., faster flight necessitates smaller deflections to achieve the same effect). Additionally, $\boldsymbol{u}$ is constrained within feasible limits by restricting it to the minimal and maximal actuator positions.

\begin{algorithm}
\caption{Calculating the control command $\boldsymbol{u}$}
\label{alg:co}
\begin{algorithmic}[1]
\State $\boldsymbol{p}_{\text{ref}} = \boldsymbol{M} \boldsymbol{\omega}_{\text{ref}}$ \;
\State $\boldsymbol{p}_{\text{meas}} = \boldsymbol{M} \boldsymbol{\omega}_{\text{meas}}$ \;
\For{$i=1 \leq 7$ (all control surfaces)}
    \State $u_i = \text{F}_i \cdot p_{\text{ref}} + \text{PID}_i(p_{\text{ref}} - p_{\text{meas}})$ \;
    \State $u_i \leftarrow u_i \frac{v^2}{v^2_{\text{ref}}}$ \;
    \State $u_i \leftarrow \text{crop}(u_i, \text{min}_i, \text{max}_i)$ \;
\EndFor
\end{algorithmic}
\end{algorithm}

In this work, we specifically focus on the development of a body rate controller to accommodate different cascaded control structures. To test the algorithm, a standard cascaded \ac{PID} controller is used as a high-level controller  for hovering on the spot against airflow:
A position control loop provides the desired orientation, which in turn determines the desired body rate.

It is worth highlighting that standard cascaded control strategies operate directly on the state (e.g., body rate) before using the mapping matrix to determine the control input for each actuator~\cite {sattar2020roll}. In contrast, our approach reverses this sequence, initially projecting the body rate onto each actuator and subsequently applying the control scheme. This methodology allows tailored tuning for each actuator, accommodating specific requirements such as the need for lower $D$-gains in instances like the inertia-heavy wing sweep compared to wing twist.

The method described in this work was first implemented and tested in simulation~\cite{song2021flightmare} before deployment on a real drone.

\subsection{Bayesian optimization}\label{sec:bo}
Multiple actuator configurations of the \textit{LisEagle} (\autoref{fig:method}e) ensure stable flight while directly influencing the lift-to-drag ratio and hence the energy consumption of the \ac{AID}. This creates a null space~\cite{dietrich2015overview} that can be explored to investigate energy-efficient configurations. Evaluating these configurations requires real-world flight to exclude model inaccuracies, making it crucial to explore this null space efficiently despite sensor noise. This leads to a multidimensional, non-convex optimization problem of identifying the configuration that results in the lowest energy consumption. \ac{BO} emerges as the preferred method due to its high sample efficiency, ability to tolerate stochastic noise, and construction of a continuous surrogate for the objective~\cite{frazier2018tutorial}, linking configurations to their resulting energy consumption.

To leverage these capabilities, the \textit{LisEagle} was flown in various configurations while measuring the corresponding energy consumption using an onboard voltage and current sensor, Holybro PM02~\cite{holybro}. The samples were obtained at \qtyproduct{5}{Hz} and averaged over \qtyproduct{10}{s}. The measurement was centred on zero by subtracting the energy consumption $e_{\text{init}}$ of the initial configuration ($c_{\text{tail}}, c_{\text{sweep}}, c_{\text{twist}}$ = 0) from all subsequent measurements. Furthermore, the noise was assumed to be homoscedastic, and the noise level of the underlying Gaussian process was adapted to match the variance $\sigma_\text{init}^2$ of the initial energy consumption measurement. For our experiments, an implementation of the BayesianOptimization library~\cite{BO} was used.
The predicted configuration was added to the control command generated by the body rate controller (\autoref{fig:method}e).
To prevent slow overwriting of the new centre positions by the integrator, the $I$ gains of the affected actuators (wing sweep, twist, and tail sweep) were set to zero. The final control command was then cropped to remain within the feasible range of the actuators.
Certain configurations may result in unstable flight (angular velocities $ > $ \qtyproduct{0.5}{rad/s} or position error $ > $ \qtyproduct{13}{cm}). These configurations were automatically marked as having high energy consumption ($e_{\text{init}} + \sigma_\text{init}$) to discourage the \ac{BO} from revisiting them. In that case, the actuators were reverted to their initial positions, and a saved set of gains was loaded to ensure stable flight; then, the search resumed.

\begin{figure}
    \centering
    \includegraphics[width=0.6\textwidth]{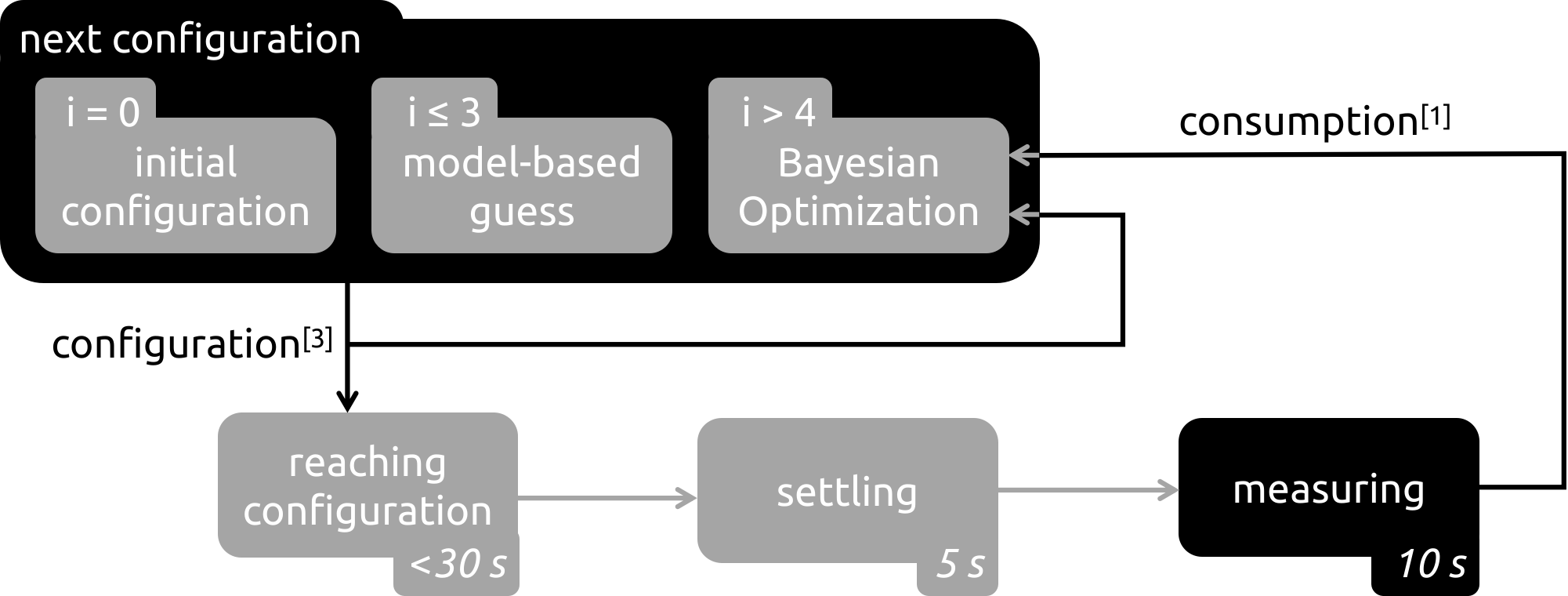}
    \caption{\textbf{Energy-efficient configurations are identified iteratively.} During the first iteration, the energy consumption of the \textit{initial configuration} is measured. The next three configurations are \textit{model-based guesses} to collect initial data for the Bayesian optimization. All subsequent configurations are determined through Bayesian optimization. Every time a new configuration is determined, it is gradually added to the control commands to allow the body-rate controller to adapt, until said \textit{configuration is reached}. 
Subsequently, the system  \textit{settles} for \qtyproduct{5}{s} before measuring and averaging the energy consumption over \qtyproduct{10}{s}. The configuration and its energy consumption are then fed into the Bayesian optimization algorithm, and the cycle repeats.}
    \label{fig:bo}
\end{figure}

Note that the \ac{BO} algorithm requires an initialization phase, in which the initial data is collected to improve the surrogate function, which in turn is utilized by the acquisition function (expected improvement with exploration parameter $\xi = 0.05$) to provide meaningful suggestions for where to sample next. In practice, this initialization is often performed through randomization or, if available, using prior knowledge of the problem. In our case, the dynamics model is leveraged by evaluating it in different configurations that ensure stable flight (low linear and angular accelerations) and high energetic efficiency (low thrust)allowing. These model-based guesses are used for the first three iterations, after which all subsequent configurations are determined by the \ac{BO} algorithm (\autoref{fig:bo}).


\backmatter

\section{Acknowledgment}
We thank Hoang Vu Phan for help in building and repairing the LisEagle as well as proofreading the manuscript.
We also thank Marc Rauch for insightful discussions on Bayesian Optimization and Raphael Zufferey for advice on experimental procedures.
This research was partly funded by the European Commission through the project Aerial Core nr. 871479 and by Armasuisse through project nr. 8003529756. Funders played no role in study design, data collection, analysis and interpretation of data, or the writing of this manuscript.

\section{Data availability}\label{sec5}
We provide videos containing different views, the corresponding flight data for each experiment and the code used to produce the findings of this study \cite{zenodo2024}.

\subsection{Take Off}\label{video:takeoff}
\subsection{Perturbation}\label{video:perturbation}
\subsection{Turbulence}\label{video:turbulence}
\subsection{Actuator 8m/s}\label{video:actuator8}
\subsection{Actuator 10m/s}\label{video:actuator10}
\subsection{Actuator 12m/s}\label{video:actuator12}
\subsection{Energy 8m/s}\label{video:energy8}
\subsection{Energy 10m/s}\label{video:energy10}
\subsection{Energy 12m/s}\label{video:energy12}
\subsection{Summary}\label{video:supercut}

\section{Declarations}
\subsection{Competing interests}
All authors declare no financial or non-financial competing interests.

\subsection{Contributions}
SJ conceived the idea and experimental design.
SJ designed and implemented the control and optimization architecture, supported by CT.
SJ, VW extended the dynamics model.
SJ and VW designed and conducted the experiments as well as the data analysis.
DF directed, supervised and fully revised the research.
All authors reviewed the manuscript.

\begin{appendices}

\section{Challenges of avian-informed drone control}\label{app1}
\begin{figure}
    \centering
    \includegraphics[width=\textwidth]{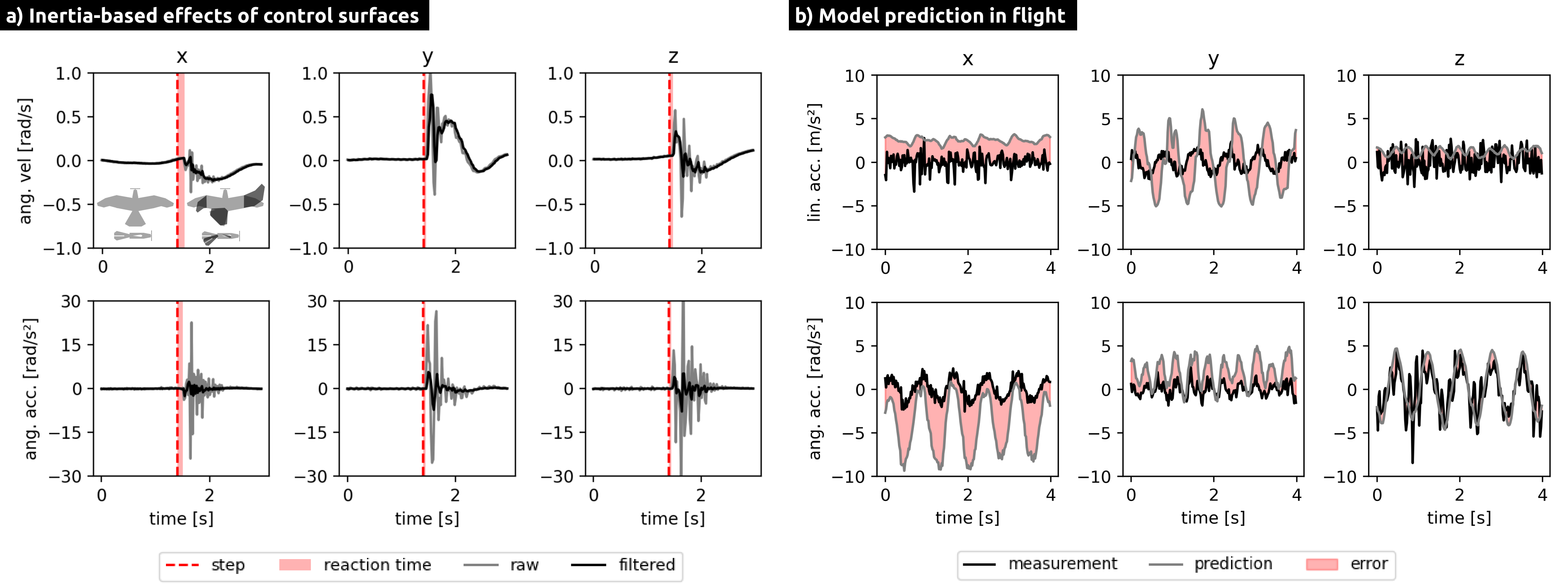}
\caption{\textbf{Experiments highlight the challenges of avian-informed drone control.} \textbf{a} Influence of actuator inertia on the angular velocity and acceleration (through differentiation) during step-like movements (red dotted line) of all actuators (mean to maximum position) with no airflow present. Tail and wing sweep actuation induce momentary angular accelerations around the $y$- and $z$-axes. The resulting angular velocities in the $x$-axis direction are slower and a byproduct of the \textit{LisEagle} rotation after the shift in the centre of gravity. The raw sensor measurements are passed through a low-pass filter. \textbf{b} Discrepancy between model predictions and measured accelerations during oscillating horizontal flight at \qtyproduct{8}{m/s}.}
    \label{fig:problem}
\end{figure}

The actuation of the morphing tail and wings, which have relatively large masses, introduces momentary torques due to the conservation of angular momentum, which is unrelated to aerodynamic effects. These inertia-based influences alter acceleration and velocity sensor readings, posing a challenge for control algorithms.
To assess the inertia effects of the control surfaces, the drone was suspended from a cable at its centre of gravity. Each aerial surface (elevator, rudder, tail sweep, left and right wing sweep, aallowingnd left and right wing twist) was actuated in a single step from the central position to the maximum range of motion, with no airflow present.
We observe angular velocities surpassing \qtyproduct{1}{rad/s} (\autoref{fig:problem}a), suggesting that momentary inertia-based effects are stronger than slower aerodynamic effects (\autoref{fig:smoke}).
To address this issue, a low-pass filter~\cite{karki2000active} is applied to the measurement data, filtering out momentary effects at the cost of introducing delay. The resulting trade-off yields a smoother, slightly delayed angular velocity signal that predominantly reflects lasting aerodynamic effects, thus enabling body-rate control. However, despite filtering, inertia-based effects remain dominant in the acceleration data, rendering acceleration-based controllers such as \ac{INDI} impractical without additional modelling refinement.

Modelling \acp{AID} is notoriously difficult. The quality of the proposed model is assessed by flying the \textit{LisEagle} in oscillatory motion at a constant velocity of \qtyproduct{8}{m/s} and comparing the collected acceleration data with the model predictions. Large discrepancies are observed (\autoref{fig:problem}b) due to unmodelled dynamic effects such as the movement of inertia-heavy actuators, aeroelasticity, and changes in airflow through the feathers. Asymmetries in the fabrication process and constraints on the \textit{LisEagle}'s position and orientation onto the force and torque sensor during the data collection process for the modelling cause constant offsets. The latter has a considerable influence on the dynamic aerodynamic coefficients. Nonetheless, our method is robust to these modelling discrepancies, as shown in the Results section.

\end{appendices}


\bibliography{sn-bibliography}

\end{document}